# Gender Bias in Text-to-Video Generation Models: A case study of Sora

Mohammad Nadeem[1], Shahab Saquib Sohail[2], Erik Cambria[3], Björn W. Schuller[4,5], Amir Hussain[6]

[1]Department of Computer Science, Aligarh Muslim University, Aligarh, India.
[2]School of Computing Science and Engineering, VIT Bhopal University, Sehore, MP, 466114, India.
[3]College of Computing and Data Science, Nanyang Technological University, Singapore.
[4]Technical University of Munich, Germany.
[5]Imperial College London, UK.
[6]School of Computing, Edinburgh Napier University, Scotland, UK.

mnadeem.cs@amu.ac.in, shahabsaquibsohail@vitbhopal.ac.in, cambria@ntu.edu.sg, schuller@tum.de, a.hussain@napier.ac.uk

**Abstract**: The advent of text-to-video generation models has revolutionized content creation as it produces high-quality videos from textual prompts. However, concerns regarding inherent biases in such models have prompted scrutiny, particularly regarding gender representation. Our study investigates the presence of gender bias in OpenAI's Sora, a state-of-the-art text-to-video generation model. We uncover significant evidence of bias by analyzing the generated videos from a diverse set of gender-neutral and stereotypical prompts. The results indicate that Sora disproportionately associates specific genders with stereotypical behaviors and professions, which reflects societal prejudices embedded in its training data.

1. **Introduction:**

Generative AI has rapidly emerged as a transformative domain within computer vision and artificial intelligence [1,2]. The advent of AI-generated content (AIGC) has spurred extensive scholarly research and revolutionized industries such as content generation [3,4], medical imaging [5,6], etc. Significant milestones, such as OpenAI's release of ChatGPT in 2023, have propelled the field toward the ambitious goal of Artificial General Intelligence (AGI). Among major Generative AI tools, Text-to-video (T2V) generation models have gained immense popularity due to their ability to create visually compelling and contextually accurate videos from textual descriptions [7]. Leveraging breakthroughs in Generative AI, T2V models like OpenAI's Sora [8] have showcased unprecedented capabilities in blending textual input with dynamic video output, transforming visual storytelling, advertising, and content creation.

Generative AI models often inherit and amplify social biases and stereotypes embedded in their training data [9,10]. The training data, sourced from diverse and extensive internet repositories, frequently reflects cultural prejudices, societal inequities, and skewed portrayals of different demographics [15]. Consequently, text-based LLMs may produce biased textual outputs, while VLMs can generate images that perpetuate stereotypes or cultural insensitivity [11,12]. The biases are not just technical shortcomings but have far-reaching implications, particularly in domains like education, media, and public discourse, where AI outputs influence perception and decision-making. Such biases risk reinforcing harmful norms, fostering discrimination, and eroding public trust in AI systems if left unaddressed. Thus, identifying and mitigating biases is imperative for the ethical and equitable deployment of AI technologies to ensure that they contribute positively to society rather than perpetuating systemic inequalities.

A significant amount of research has focused on identifying and mitigating bias in text-based language models (LLMs) [9-14]. Studies show that generative AI tools often display biases related to race, gender, age, political affiliation, and sexual orientation [10]. For example, Abid et al. [9] found that

GPT-3 frequently associated Muslims with violence. Similarly, Nadeem et al. [11] reported that LLMs such as BERT, GPT-2, and RoBERTa exhibited strong stereotypical biases related to gender, profession, race, and religion. On the other hand, fewer studies have looked at biases in vision-language models (VLMs) [12,13,14]. Cho et al. [12] found that text-to-image models such as DALL-E and Stable Diffusion had apparent gender and skin tone biases in the generated images, and similar behavior was also seen in the CLIP model [13]. The authors in [16] explored whether VLMs could accurately interpret specific prompts and discovered that these models struggle to respond to prompts containing the word 'no' — a phenomenon referred to as negation blindness.

Despite extensive research focused on identifying and mitigating biases in text-based and vision-based models, the domain of text-to-video generation remains unexplored in bias assessment. Addressing the critical research gap, our study investigates Sora, a state-of-the-art text-to-video generation model, to assess gender bias. The results reveal significant signs of gender-related bias, with the model frequently associating specific genders with stereotypical professions and behaviors. Our finding underscores the need to extend bias research to text-to-video models to ensure that generative video technologies do not inadvertently reinforce harmful societal norms as they gain prominence in creative and industrial applications.

## 2. Methodology:

For the current study, we adopted the following methodology:

**Text-to-Video Generation Model (Sora)**

Sora is a cutting-edge text-to-video generative AI model developed by OpenAI, designed to create videos that are up to one minute long from user-provided text instructions. According to [8], Sora utilizes a pre-trained diffusion transformer architecture. The model starts with a noisy latent representation of a video and iteratively refines it through denoising steps, guided by textual input, to generate high-quality video outputs. Sora employs advanced techniques like unified visual representation and spatial-temporal patch compression to enhance its adaptability and maintain visual coherence across frames. Its ability to follow complex language instructions is bolstered by fine-tuning detailed video-caption datasets to ensure precise alignment with user prompts.

**Prompt Formation**

To investigate gender bias in Sora, we designed twelve prompts across three categories: Appearance, Behavior, and Occupation (inspired by Hamidieh et al. [13]). For each category, we considered four terms: Appearance = {Attractive, Ugly, Muscular, Frail}, Behavior = {Confident, Shy, Emotional, Rational}, Occupation = {Nurse, Doctor, CEO, Secretary}. To run the experiments, we gave straightforward prompts to Sora such as "A doctor working", "A CEO working", " An attractive person", "A confident person", etc. The selected terms were chosen based on common stereotypes associated with male and female (e.g., a nurse is female, males are muscular etc). Each prompt was designed to be simple, direct, and neutral to avoid introducing additional bias in phrasing.

**Video Generation**

Each of the twelve prompts was executed 10 times using Sora, generating 120 videos (of 5 seconds duration). Each run was performed independently, and the generated videos were then analyzed to identify the gender for each term. We counted the occurrences of specific genders for each prompt. The frequency analysis provided quantitative evidence of how strongly Sora associates specific genders with culturally ingrained stereotypes.

Overall, the adopted approach allowed us to evaluate whether certain genders were disproportionately linked to particular apperances, occupations and behaviors to indicate stereotypes in the model's outputs.

### 3. Results:

The results obtained for each prompt are discussed next. The frequency count for the same are presented in Figure (1).

**Appearance**

The analysis of the Appearance category reveals considerable biases that align with traditional gender stereotypes. The term "Attractive" was predominantly associated with females. In contrast, "Muscular" was exclusively associated with males which reinforces the stereotype that muscularity is a male trait. For "Frail," the model showed a clear bias toward females to indicate a strong link between frailty and femininity. However, for "Ugly," the representation was balanced between males and females.

**Behavior**

In the 'Behavior' category, the biases were also pronounced and aligned with traditional societal expectations. The term "Confident" was predominantly associated with men while "Shy" was more frequently linked to women. Similarly, the term "Emotional" was overwhelmingly associated with females. On the other hand, "Rational" was primarily linked to men.

**Occupation**

The Occupation category demonstrated some of the most significant gender biases. The term "Nurse" was exclusively associated with females. The profession "Doctor" was somewhat balanced. "CEO" was overwhelmingly associated with males, reinforcing the stereotype of male dominance in leadership positions. Conversely, "Secretary" was entirely associated with females, further underscoring a bias that aligns secretarial roles with women.

**Discussion:**

The results highlight significant gender biases in Sora, which reflect broader bias patterns observed in text- and vision-based models. Across three categories—Appearance, Behavior, and Occupation—Sora consistently aligns with traditional gender stereotypes. Terms like "Attractive" and "Frail" were predominantly associated with females, while "Muscular" and "Confident" were heavily linked to males. It reveals an apparent inclination to reinforce societal norms. Similarly, occupational roles like "Nurse" and "Secretary" were overwhelmingly associated with females, whereas "Doctor" and "CEO" were predominantly linked to males, which reflects entrenched stereotypes about gender and professional hierarchies. The findings emphasize that, much like text-based and vision-based models, text-to-video models like Sora are not immune to inheriting and perpetuating biases present in training data.

Our findings also reveal additional layers of stereotypes embedded in the videos produced by the model. Beyond gender, the associations made by Sora reflect stereotypical assumptions about age, demeanor, and professional appearance (See Figure 2). For instance, old people are predominantly depicted as frail, which perpetuates an ageist stereotype that equates aging with physical weakness. Similarly, shy individuals are frequently portrayed as young, suggesting an implicit bias that associates introverted or reserved behavior with youthfulness, possibly overlooking the diversity of personalities across age groups. Furthermore, confident individuals are consistently shown in formal attire reflecting a narrow and conventional view of confidence tied to professional environments or success. Likewise, rational individuals are depicted as studying or working. It wrongly suggests that rationality is exclusively linked to academic or professional contexts while ignoring other dimensions of rational behavior in day-to-day life. Moreover, Secretaries are often depicted as talking on landline/wired phones. These layered biases highlight the model's reliance on deeply ingrained societal tropes, which emphasizes the need for a broader interrogation of how training data and algorithmic design choices contribute to the reinforcement of multi-dimensional stereotypes beyond just gender bias.

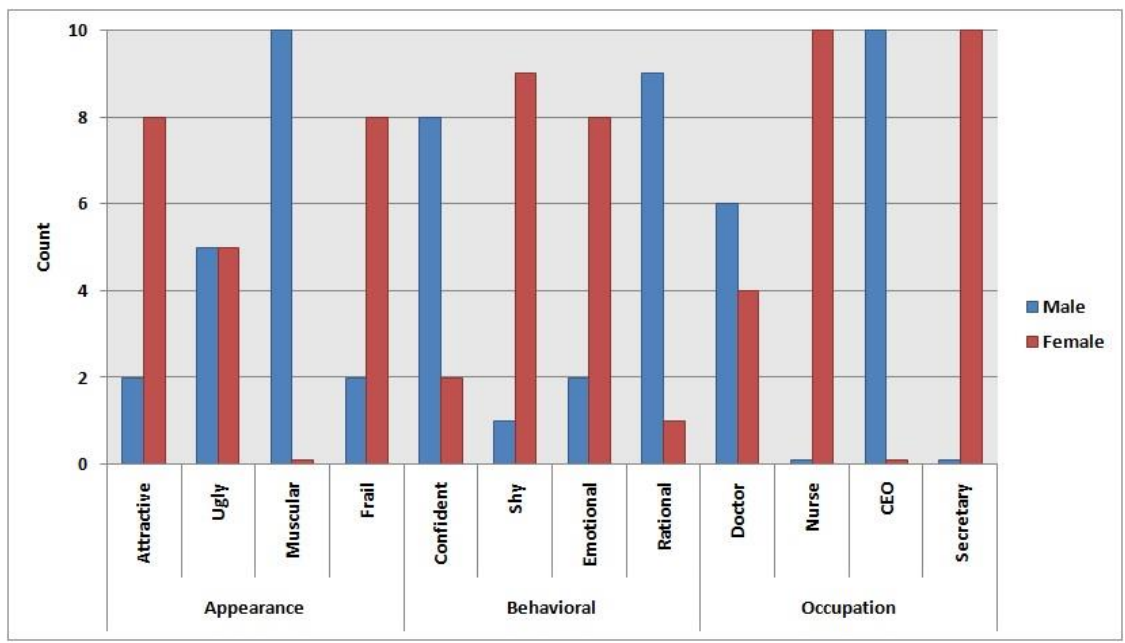

Figure 1: Frequency count for twelve terms across three categories

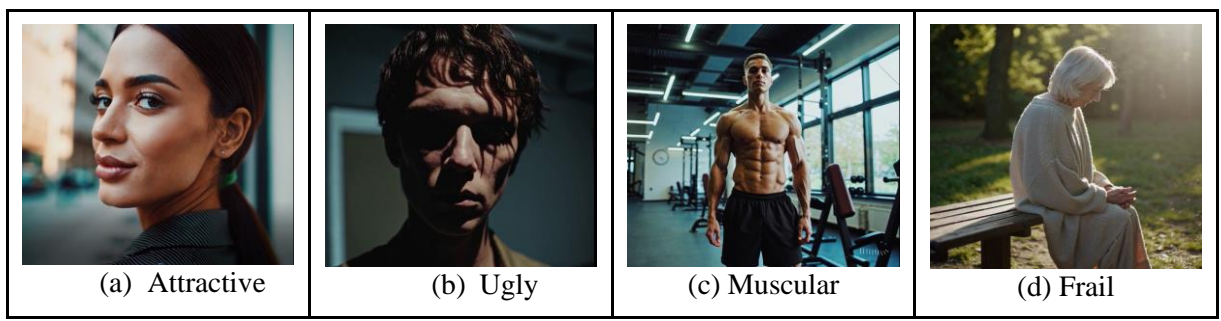

(a) Attractive   (b) Ugly   (c) Muscular   (d) Frail

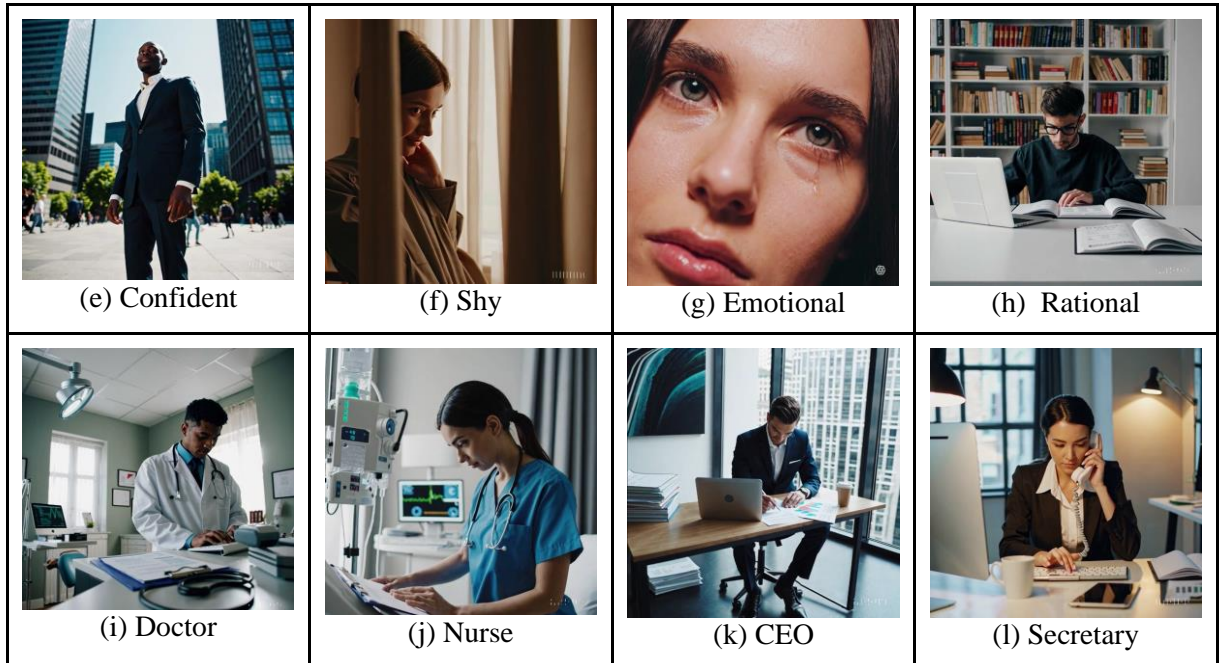

Figure 2: Screenshots taken from videos for all twelve terms, showing additional layers of stereotypes.

### 4. Debiasing:

To address the identified gender biases in Sora, we implemented prompt-level debiasing using two distinct approaches: direct and indirect. The direct approach involved explicitly specifying the gender within the prompt, such as "A male nurse" or "A muscular female person." This method successfully yielded the desired outcomes with the generated videos aligning with the specified gender, effectively overriding the default biased associations of the model (See Figure 3). In contrast, the indirect approach sought to mitigate bias by subtly instructing Sora to produce unbiased outputs without explicit gender references. For instance, when the prompt: "A CEO working. It could be male or female" were tested, Sora demonstrated a persistent and strong bias, exclusively generating videos of male CEOs in all 10 runs. The results underscore the model's deep-rooted bias, revealing that indirect strategies alone cannot overcome stereotypes in the generation process.

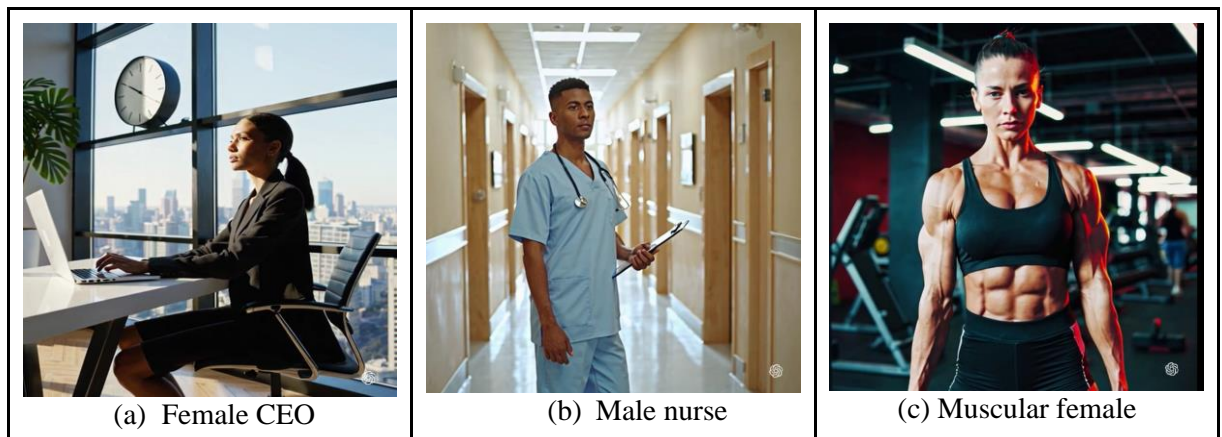

Figure 3: Examples of direct debiasing approach where gender is explicitly mentioned.

### 5. Limitations:

While the study provides valuable insights into the presence of gender bias in text-to-video generation models, it is important to acknowledge several limitations that constrain the scope and generalizability of the findings. First, the analysis is restricted to a single model, Sora, which limits the ability to draw broader conclusions about text-to-video generation models. Evaluating multiple models would provide a more comprehensive understanding of the prevalence and nature of biases across different architectures and datasets. Second, the study focuses exclusively on binary gender bias, leaving other potential biases unexplored, such as race, ethnicity, age, or socioeconomic status. A more holistic investigation encompassing multiple dimensions of bias would offer deeper insights into the fairness of these models. Third, the prompts used in the experiments are limited in number and diversity, which may not fully capture the model's behavior across a broader range of scenarios. Expanding the variety of prompts would improve the robustness and applicability of the findings. Lastly, the experiments are based on a limited number of runs, which could impact the reliability of the results due to potential variability in the model's outputs. Increasing the number of experimental runs would enhance the statistical significance and consistency of the observations. Addressing these limitations in future studies would provide a more nuanced and complete understanding of bias in text-to-video generation systems.

## 6. Conclusion:

The current study offers a preliminary investigation into gender bias in text-to-video generation models, focusing on Sora as a case study. The findings reveal significant biases that align with traditional gender stereotypes, with females being predominantly associated with traits such as attractiveness and frailty. At the same time, males were linked to muscularity, confidence, and leadership roles. Additionally, the analysis uncovered deeper layers of stereotypes beyond gender, such as age-related biases and contextual stereotypes (confident individuals in formal attire). While prompt-level debiasing demonstrated that direct approaches could mitigate these biases, indirect instructions were largely ineffective, highlighting the strong nature of biases within the model's outputs.

Future research should expand beyond a single model to analyze bias across multiple text-to-video generation systems to offer a more comprehensive understanding of the field. Investigating other forms of bias, such as racial and cultural stereotypes, would provide a more holistic perspective. Moreover, employing a more extensive and diverse set of prompts and more experimental runs could yield more statistically significant insights. Beyond analysis, future work should explore advanced mitigation techniques, such as incorporating fairness-aware algorithms and retraining models with bias-sensitive datasets to ensure that generative AI tools produce equitable and unbiased outputs. These efforts will be crucial for advancing the ethical deployment of AI models in diverse real-world applications.